\documentclass[conference]{IEEEtran}
\IEEEoverridecommandlockouts
\usepackage{cite}
\usepackage{amsmath,amssymb,amsfonts}
\usepackage{algorithmic}
\usepackage{graphicx}
\usepackage{textcomp}
\usepackage{xcolor}
\usepackage{gensymb}
\usepackage{mathpazo}
\usepackage{algorithm}
\usepackage{esvect}
\usepackage{amsmath}
\usepackage{tikz}
\usepackage{subfig}

\def\BibTeX{{\rm B\kern-.05em{\sc i\kern-.025em b}\kern-.08em
    T\kern-.1667em\lower.7ex\hbox{E}\kern-.125emX}}

\begin{document}
\pagenumbering{gobble}
\begin{center}
\textbf{This work has been submitted to the IEEE for possible publication. Copyright may be transferred without notice, after which this version may no longer be accessible.}
\end{center}
Begin main text on the second page
\newpage

\title{Vehicle State Estimation through Modular Factor Graph-based Fusion of Multiple Sensors}

\author{Pragyan Dahal$^{1}$, Jai Prakash$^{1}$, Stefano Arrigoni$^{1}$, Francesco Braghin$^{1}$% <-this % stops a space
\thanks{$^{1}$ Dept. of Mechanical Engineering, Politecnico di Milano, Italy}}

\maketitle
\thispagestyle{plain}
\pagestyle{plain}
 
\begin{abstract}

This study focuses on the critical aspect of robust state estimation for the safe navigation of an Autonomous Vehicle (AV). Existing literature primarily employs two prevalent techniques for state estimation, namely filtering-based and graph-based approaches. Factor Graph (FG) is a graph-based approach, constructed using Values and Factors for Maximum Aposteriori (MAP) estimation, that offers a modular architecture that facilitates the integration of inputs from diverse sensors. However, most FG-based architectures in current use require explicit knowledge of sensor parameters and are designed for single setups. To address these limitations, this research introduces a novel plug-and-play FG-based state estimator capable of operating without predefined sensor parameters. This estimator is suitable for deployment in multiple sensor setups, offering convenience and providing comprehensive state estimation at a high frequency, including mean and covariances.
The proposed algorithm undergoes rigorous validation using various sensor setups on two different vehicles: a quadricycle and a shuttle bus. The algorithm provides accurate and robust state estimation across diverse scenarios, even when faced with degraded Global Navigation Satellite System (GNSS) measurements or complete outages. These findings highlight the efficacy and reliability of the algorithm in real-world AV applications.
\end{abstract}


\section{Introduction}

The precise estimation of the state of an ego vehicle is a fundamental requirement for both ADAS systems and Autonomous Driving. It involves obtaining accurate information about the vehicle's location, orientation, velocity, and other state components, enabling informed decision-making in subsequent software components. Various functionalities such as vehicle navigation, obstacle state estimation \cite{ObstacleState}, and path planning heavily rely on reliable and robust vehicle state information. Similarly, for effective vehicle control, a comprehensive understanding of the vehicle's state is crucial.

While Global Navigation Satellite Systems (GNSS) are commonly employed for vehicle localization \cite{msf}, \cite{Integrated_Ego_Estimation}, their reliability can be compromised in urban environments due to factors like limited satellite visibility caused by high-rise buildings and other obstructions. The accuracy of GNSS measurements can also vary based on the availability of Real Time Kinematic (RTK) corrections. To mitigate these challenges, sensor fusion techniques have been explored in the literature. By integrating GNSS with other proprioceptive and exteroceptive sensors, such as inertial measurement unit (IMU), steer encoder and wheel speed encoders \cite{vehicle_dynamics_pre}, Lidar\cite{liosam}, and Camera, state estimation redundancy and reliability can be enhanced. Sensor fusion also improves the algorithm's robustness to GNSS accuracy degradation and outages.

In this work, we propose a state estimator that leverages the modular architecture of factor graphs for Maximum a Posteriori (MAP) estimation through sensor fusion. By utilizing the available set of sensors, our approach provides a comprehensive state estimation solution. The hybrid architecture incorporates filters for predictions while employing optimization techniques for real-time state estimation at a high frequency. This allows for accurate and timely decision-making in ADAS systems and autonomous driving scenarios.

\begin{figure}[t]
\centering
\includegraphics[width=9cm,height=4.5cm]{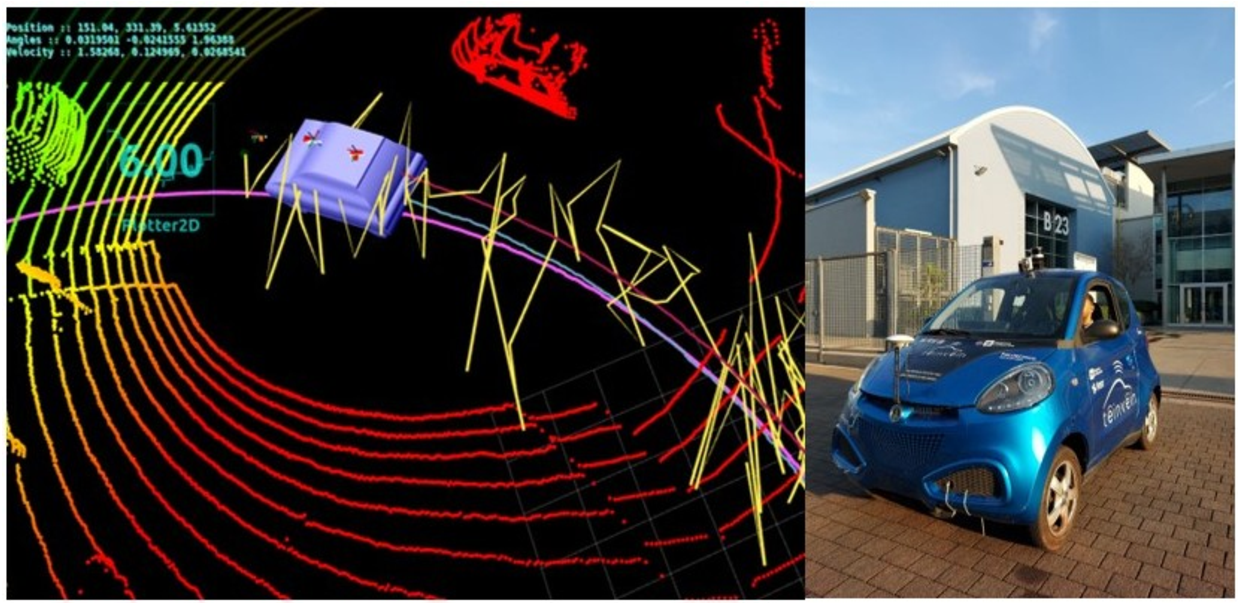}
\caption{An instance of vehicle state estimation using the proposed algorithm in Setup 2 for quadricycle vehicle (shown in the right). Trajectories labels: Yellow:-GNSS measurements, Pink:- FG estimates for Setup 1, Blue:- FG estimates for Setup 2 }
\label{fig::fig1}
\end{figure}

The contributions of the paper are:
\begin{itemize}
    \item We present a lightly coupled factor graph-based modular architecture to estimate the complete state at real time with high frequency up to $100 Hz$ of an Autonomous Vehicle using GNSS, Lidar, IMU, and Vehicular Sensors.
    \item We perform a comparative study of the proposed architecture with the traditional Extended Kalman Filter(EKF) \cite{jai2023_2} and with Ground Truth Data obtained with the use of the RTK-corrected GNSS sensors.
\end{itemize}

The remainder of the paper is organized in the following fashion. Section \ref{SoA} presents a survey of the related works on the State Estimation. Section \ref{Problem} explains the objective and problem formulation for this work. In Section \ref{FG_Construction}, we introduce the Factor Graph construction and sensor setups used in this work. In Section \ref{Experiments}, we discuss the experimental setup and validate the proposed architecture for different sensor setups. Section \ref{Conclusion} concludes the paper.  

\section{Related Works}\label{SoA}
Ego Vehicle State Estimation for ADAS systems and Autonomous Driving employs various sensors, including GNSS, IMU, LIDAR, and cameras. These estimation approaches can be broadly categorized into Filtering-Based and Graph-Based methods.

Filtering algorithms such as Extended Kalman Filter (EKF) \cite{jai2023_2}, Unscented Kalman Filter (UKF) \cite{Integrated_Ego_Estimation}, and Particle Filter (PF) have been widely utilized due to their simplicity. For example, Prakash et al. \cite{jai2023_2} formulated an EKF with a single-track model to estimate side-slip angle, lateral tire forces, and vehicle pose. They can provide sub-optimal results and are prone to inaccuracies when dealing with highly nonlinear system models. Additionally, filtering algorithms suffer from limitations in incorporating delayed measurements and correcting past errors.

To overcome the downsides of filtering algorithms, Graph-Based optimization techniques offer a more robust solution. In Graph-Based optimization, the optimal states are determined by solving the Maximum A Posteriori (MAP) estimation over the joint distribution of the entire graph. Two common types of graph-based methods are Pose Graph (PG) and Factor Graph (FG) approaches.

Pose Graph-based methods establish relationships between successive poses and perform MAP estimates. These methods, as seen in studies such as \cite{frosi2021artslam} and \cite{kissicp}, are computationally efficient but are limited to Gaussian and linear relationships. On the other hand, Factor Graphs \cite{factor_graphs_for_robot_perception} construct graphs composed of factors and values, with factors representing constraints on or between poses. FG-based methods, widely used in motion and state estimation, offer more flexibility. For example, in VINS-Mono \cite{VINSmono}, FG is utilized to estimate ego-motion by fusing Camera and IMU data. Pre-integrated IMU measurements and Camera-generated features serve as factors in the FG to estimate Camera motion. Some studies, such as \cite{vehicle_dynamics_pre}, extend the IMU pre-integration factor by incorporating in-vehicle sensors. Other variants of FG-based algorithms, like Lio-SAM \cite{liosam}, are specifically designed for Lidar-Inertial Odometry but may be restricted to specific IMU and Lidar sensor configurations.

A multi-sensor fusion hybrid algorithm for excavators using a dual-factor graph architecture is proposed in \cite{msf}. However, the dual graph implementation is observed to be prone to failures during sensor switching in certain scenarios. The work in \cite{lv2023continuoustime} represents vehicle motion as a spline continuous in time within an FG to estimate vehicle state. Despite the advantages of FG-based approaches, achieving high-frequency implementation and modular construction with continuous-time representation remains challenging and complex.

In this study, we develop a factor graph-based state estimator inspired by the hybrid nature of \cite{msf}, incorporating filter-based prediction and optimization updates. Our approach includes additional factors and prediction steps to account for vehicle dynamics and perform covariance prediction within the filtering step. We validate the algorithm using multiple sensor setups and demonstrate its robustness to GNSS noise and outages.

\section{Problem Formulation} \label{Problem}
The primary goal of this paper is to estimate the state of the Ego Vehicle by utilizing measurements from various sensors including GNSS, LIDAR, IMU, as well as in-vehicle sensors like Wheel and Steering Encoders. The modular design of the algorithm allows for the flexible inclusion or exclusion of sensor measurements in the Factor Graph (FG) based on their availability.

In our implementation, we choose the IMU sensor location as the reference point for estimating the vehicle's state. The set of IMU states up to time instance $t$ in the Global Reference Frame (G) is denoted as $X^{G}_t$ and can be expressed as:

\begin{equation}
X^{G}_t = {x_0^G, x_1^G, ..., x_t^G}
\end{equation}

Each state at time instance $t$, denoted as $x_t^G$, consists of the 3D IMU position and angles in $G$, velocity in the IMU reference frame (I), and IMU biases. During the estimation process, the algorithm operates in the local Odometry Frame (O), and we subsequently transform the estimated state to $G$ using the known initialization transformation ${}_{0}\textbf{T}^{G}$.

\begin{equation}
x_i^G = \begin{bmatrix}p_i^G, R_i^G, v_i^I, b_i^I\end{bmatrix}
\end{equation}

In the above equation, $R_i^G \in SO(3)$ represents the orientation of the IMU in $G$. The sensors employed in this architecture operate asynchronously and provide measurements at different time instances. Therefore, based on their arrival timestamps, we construct a factor graph that is optimized to obtain the state estimates at the corresponding time instances.

\begin{figure*}[h]
\centering
\subfloat[Factor Graph and the Sensor Setup 1 for the quadricycle vehicle.]{\includegraphics[width=0.5\linewidth]{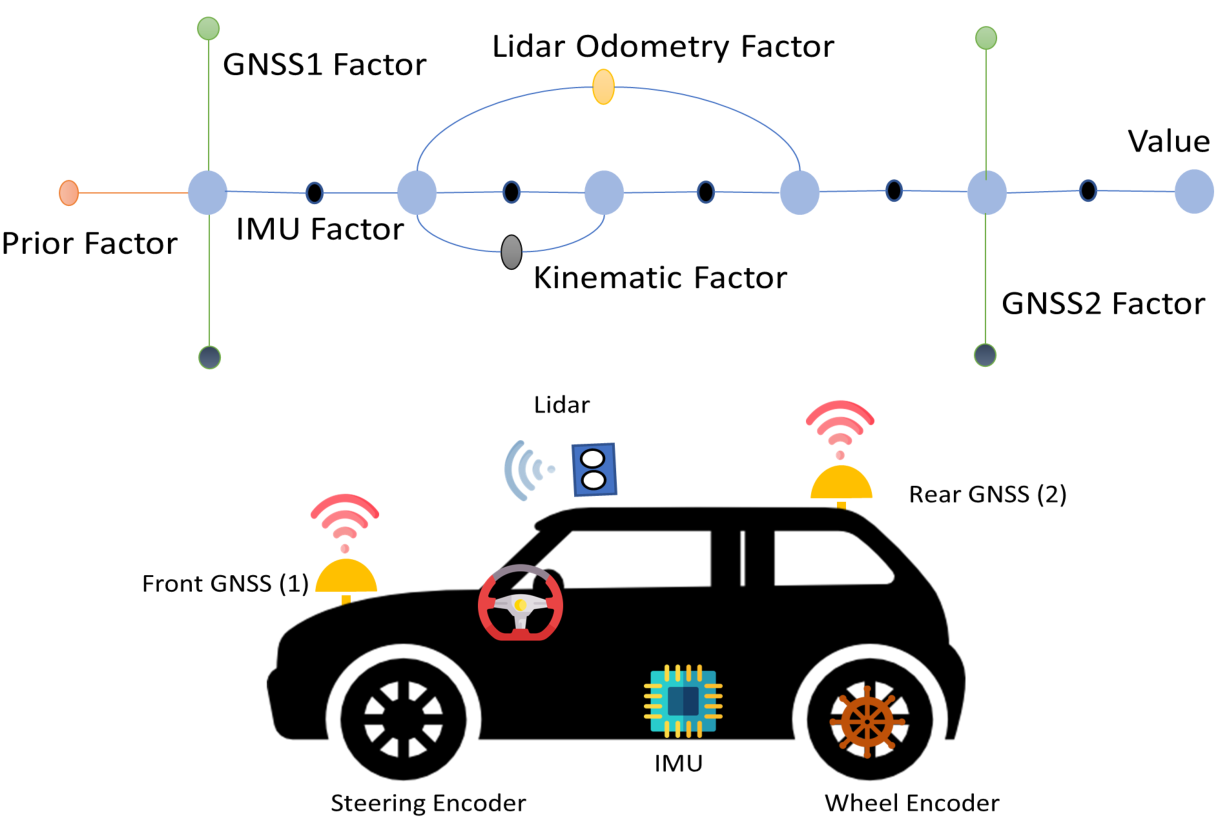}%
\label{fig:FG_Teinvein}}
\hfil
\subfloat[Factor Graph and the Sensor Setup 3 for the EasyMile Shuttle]{\includegraphics[width=0.5\linewidth]{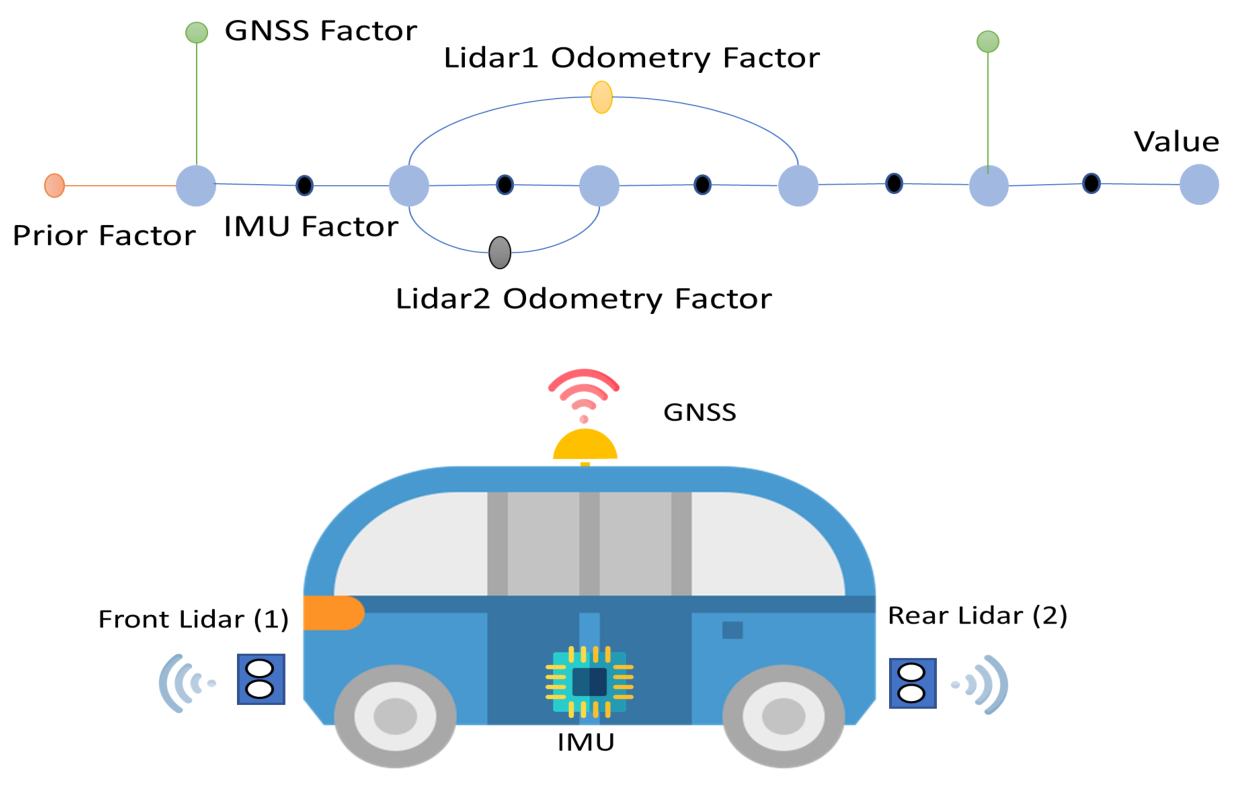}%
\label{fig:FG_Easymile}}
\caption{Factor Graph Architectures demonstrating the Modular Nature of the Algorithm and the Sensor Setups }
\label{fig:FG_Construction}
\end{figure*}

\section{Factor Graph Construction} \label{FG_Construction}
We construct our FG using GT-SAM library \cite{gtsam}. While constructing the FG, we take inspiration from the work proposed by \cite{msf} and add the Values to the Factor Graph corresponding to the timestamp of each IMU measurement. The framework is based on the Robot Operating System(ROS), which enables adding a Value in the FG when an IMU message is received in the ROS Network. Unlike \cite{msf}, where authors construct two FGs to deal with the re-orientation of the poses and GNSS sensor failure, we construct a single FG as we observed system failure when switching between two factor graphs.

\subsection{Sensor Setups} \label{section:Sensor_Setups}
We conducted our study using three different Factor Graph (FG) setups, each corresponding to a different sensor configuration. The first FG setup represents the quadricycle vehicle with front-wheel drive, as depicted in Figure \ref{fig:FG_Teinvein}. This setup includes two RTK-corrected GNSS sensors, positioned at the front and rear of the vehicle, a 32-plane Hesai Lidar mounted on the roof, an IMU sensor located at the vehicle's center of gravity, steer encoder and speed encoders. The second setup also corresponds to the quadricycle vehicle. To assess a worst-case scenario, the second setup employs only the data from the front RTK-corrected GNSS sensor while excluding the rear sensor. Additionally, artificial noise is intentionally introduced to corrupt this data. The purpose of this deliberate corruption is to simulate the expected level of positional uncertainty, corresponding to a circular error probable (CEP) of 2m.
\par
The third setup of the Easymile with all-wheel drive, illustrated in Figure \ref{fig:FG_Easymile}, features a single RTK-corrected GNSS sensor positioned at the geometric center of the vehicle in the top view. Additionally, this setup includes two 16-plane Velodyne Lidar sensors, placed at the front and rear of the vehicle, and an IMU sensor. These three setups allow us to evaluate the performance of our algorithm under different sensor configurations, providing insights into the robustness and accuracy of the state estimation process in various scenarios.

\subsection{IMU Factor}
To model the IMU measurements, we utilize the widely used pre-integrated IMU factor, as proposed in \cite{imu_preintegration}. Given the raw IMU measurements at time instance $t$, including the angular velocities ${}^I\hat{\omega}_t$ and linear acceleration ${}^I\hat{a}_t$, the accelerometer and gyroscope models can be expressed as shown in \cite{imu_preintegration}:

\begin{equation}
\begin{aligned}
    {}^I\hat{\omega}_t = {}^I\omega_t + {}^Ib_{\omega_t} + {}^I\eta_t, \\
    {}^I\hat{a}_t = {}_I^WR_t( {}^Wa_t-{}^Wg) + {}^Ib_{a_t} + {}^I\nu_t
\end{aligned}
\end{equation}

In the above equations, ${}^I\omega_t$ and ${}^Ia_t$ represent the true angular velocities and linear accelerations, respectively. The IMU angular velocity and linear acceleration biases, denoted as ${}^Ib_{\omega_t}$ and ${}^Ib_{a_t}$, are modeled as a random walk process. The noise terms, ${}^I\eta_t$ and ${}^I\nu_t$, follow Gaussian distributions. ${}_I^WR_t$ is the rotation matrix from ENU to IMU frame. For a more comprehensive explanation of the IMU factor, please refer to \cite{imu_preintegration}.

\subsection{GNSS Unary Factor}
In quadricycle vehicle setup 1, we utilize two GNSS sensors: one located at the rear of the vehicle rooftop and another at the front, as depicted in Figure \ref{fig:FG_Teinvein}. To incorporate these GNSS measurements into the state estimation process, we first transform them into the IMU positions using the known static transformation between the GNSS sensors and the IMU:

\begin{equation}
    \begin{aligned}
        {}^{I}\hat{p}_{W} = {}^{G}\hat{p}_{W} + {}_{G}^IR_t. {}^{I}_Gp_{W}
    \end{aligned}
\end{equation}

Here, ${}^{I}\hat{p}{W}$ represents a GNSS measurement transformed into the IMU location expressed in the Global Reference Frame (GRF). ${}^{G}\hat{p}{W}$ denotes the actual GNSS measurement expressed in the GRF, and ${}_{G}^IR_t$ and ${}^{I}Gp{W}$ represent the rotation and translation for the static transformation between the GNSS and IMU sensors, respectively. This transformation is performed for all the GNSS measurements. Once the GNSS measurements are transformed into the IMU location within the vehicle, they are incorporated into the Factor Graph (FG) as unary factors associated with the closest timestamp. Additionally, in the case of quadricycle vehicle first setup, since the GNSS sensors are RTK corrected and provide precise position measurements with accuracy in the centimeter range, we compute the global yaw and pitch angles based on the presence of two GNSS sensors in the car. These angles are also included as factors in the FG, along with the position measurements. To ensure the reliability of the measurements incorporated into the FG, we implement a thresholding system based on the GNSS position covariance values. This system helps discard extremely unreliable measurements, ensuring the overall quality and accuracy of the state estimation process.

\subsection{Lidar Odometry Factors}
We obtain the Lidar Odometry by implementing the simple ICP-based odometry computer proposed in \cite{kissicp}. Lidar Odometry is computed as the relative transformation between two consecutive Lidar frames. The 6DOF Lidar Odometry is transformed into the IMU position and added to the FG. The easy-to-integrate nature of KISS-ICP, \cite{kissicp} is also exploited here to make the system modular, which can work directly with the point cloud data recorded with any Lidar sensor. 

\subsection{Kinematic Factors}
A simple kinematic factor based on the vehicle kinematics is also added to the FG when the Velocity and Steering measurements are obtained from the respective sensors. The velocity encoder provides the longitudinal velocity of the vehicle while the steering encoder provides a steering angle measurement. If $V_x$ and $\delta$ are the longitudinal velocity and steering angle measured by the encoders, the integrated odometry from the vehicle kinematics using:
\begin{equation}
    \begin{aligned}
        \hat{\omega}_W = V_x\tan(\delta)/l
    \end{aligned}
    \label{ackermann}
\end{equation}
$\hat{\omega}_W$ is the yaw rate. The measured velocity and derived yaw rate are used to  compute the integration factors between the connecting values in the FG. Due to kinematic relation (eq \ref{ackermann}), the factors are computed in the GRF centered on the rear axle and moved to the IMU position using a static transformation. 

\section{Results and Comparision} \label{Experiments}
\begin{figure*}
\centering
\includegraphics[width=16cm,height=8cm]{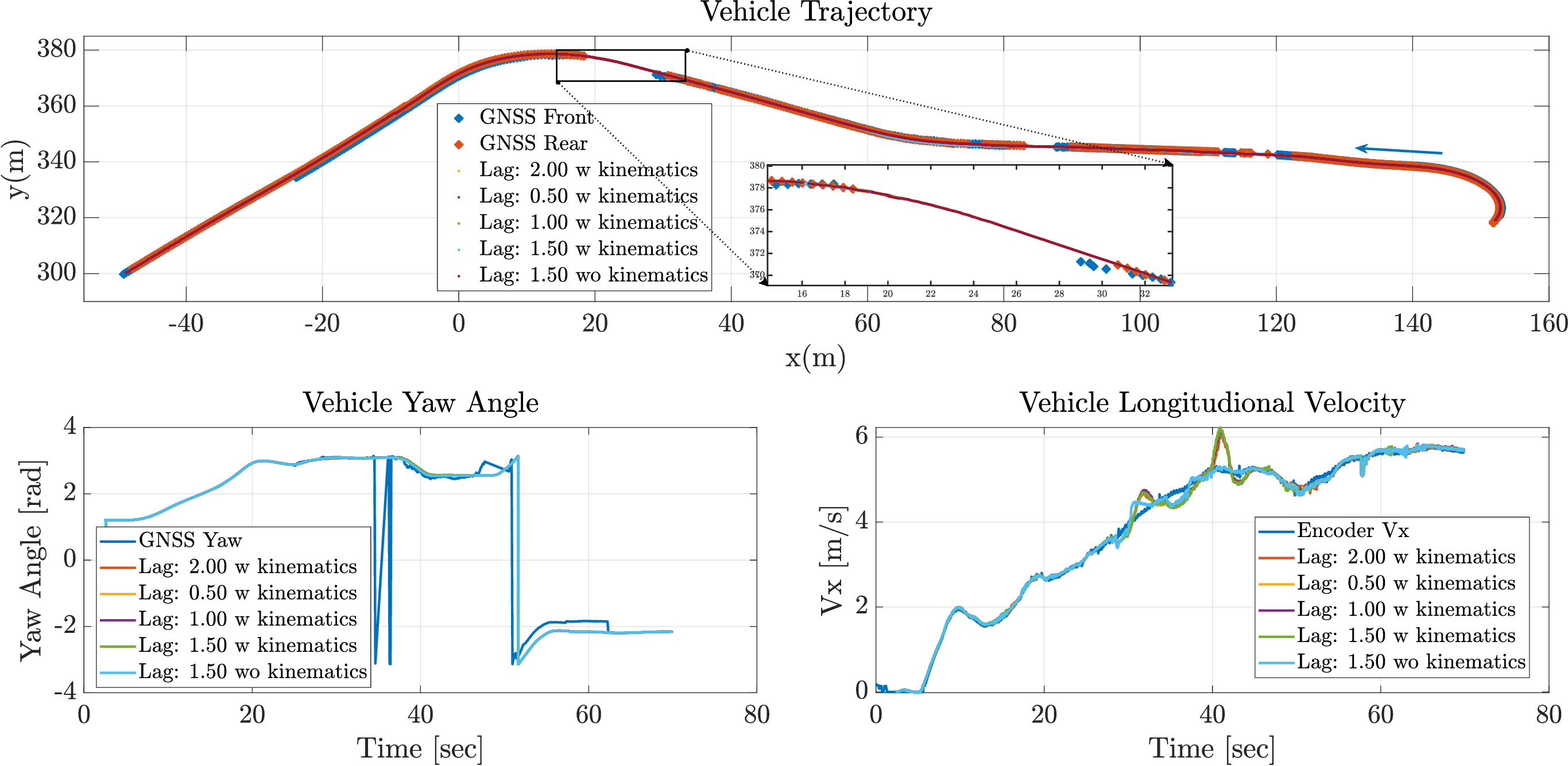}
\caption{Estimated position, yaw, and velocity for setup 1 of quadricycle for different time windows for optimization, with and without the use of kinematic factor}
\label{fig::gt_all}
\end{figure*}

In this section, we evaluate three different sensor setups, as discussed in \ref{section:Sensor_Setups}. We first utilize setup 1 to establish ground truth values for setup 2. Setup 1 comprises two RTK-corrected GNSS sensors, which can achieve centimeter-level accuracy when the RTK correction functions properly. Additionally, the positioning of these sensors in the vehicle allows for the extraction of information regarding the global heading and pitch angle.

For setup 1, we perform state estimation using the Factor Graph (FG) approach and conduct various analyses on the optimization batch time. Figure \ref{fig::gt_all} illustrates the state estimation results corresponding to different window lengths considered for optimization. We analyze window lengths ranging from 0.5 seconds to 2 seconds. Since we create values for each IMU measurement at a frequency of 100Hz, each second involves 100 values in the optimization process. We observe that the use of these different optimization windows does not significantly affect the results.

Furthermore, we analyze the results with and without the addition of the vehicle kinematic factor in the FG. Although we do not observe significant disparities due to the already good estimates from the RTK-corrected GNSS sensors, the inclusion of the kinematic factor enhances robustness in cases of sensor failure and misalignment. It is noteworthy that even without the kinematic factor, the FG-based algorithm provides longitudinal velocity estimates that resemble the measurements from the velocity encoder. The yaw estimates also exhibit consistency with the GNSS yaw values and remain robust in cases of GNSS failures. \par

\begin{figure*}
\centering
\includegraphics[width=16cm,height=8cm]{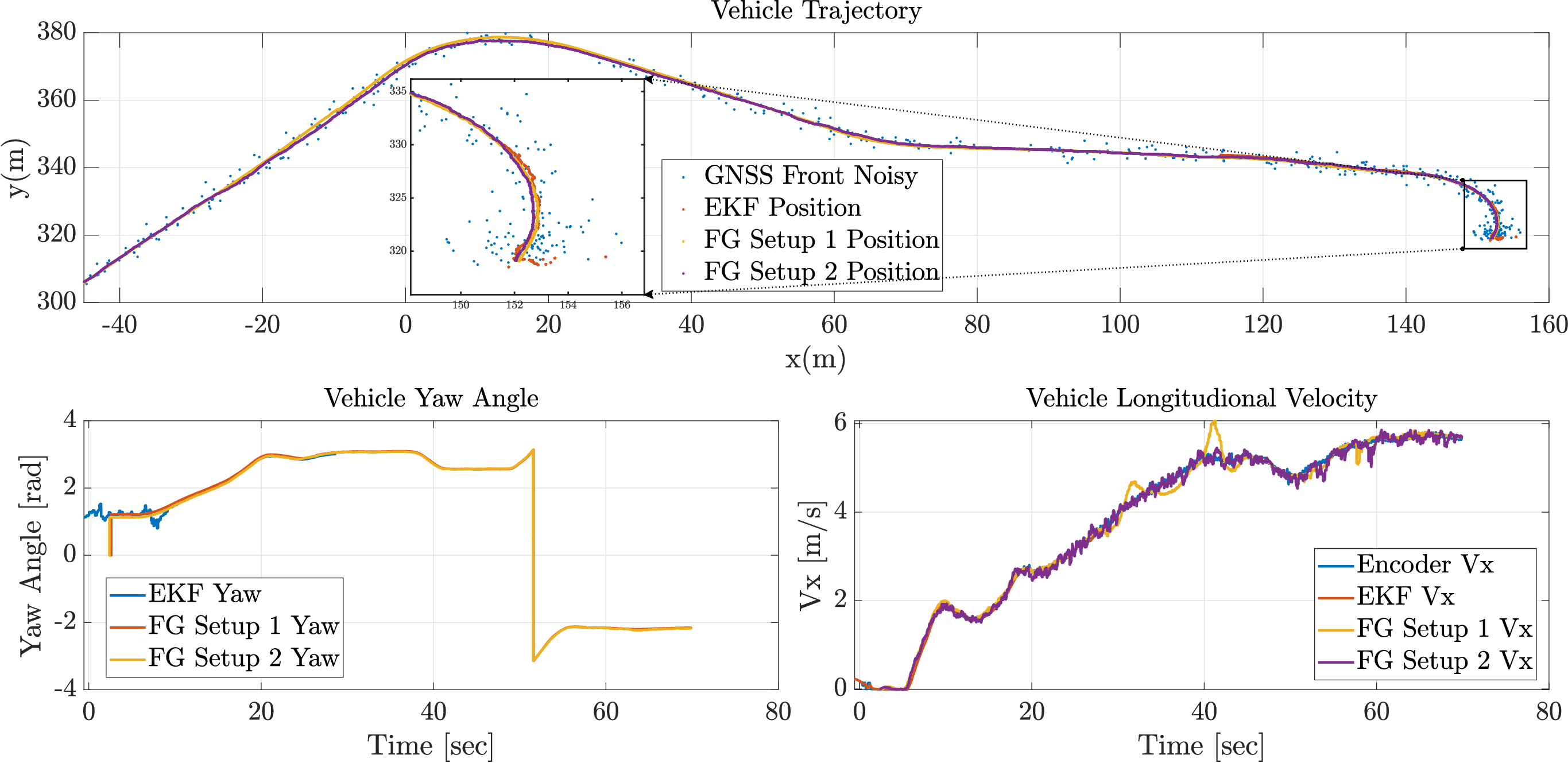}
\caption{State Estimation results for setup 2, one noisy GNSS, one Lidar, and one IMU of quadricycle car compared against the setup 1 with full sensor availability }
\label{fig::noisy_all}
\end{figure*}

For quadricycle's setup 2, consisting of one front GNSS sensor, one Lidar, and one IMU sensor, we introduce simulated noise to the GNSS sensor to create a more realistic measurement scenario. This setup allows us to evaluate the robustness and precision of the FG algorithm in cases of RTK correction failure and GNSS outages, which are common in urban environments. It also demonstrates the modular architecture of the proposed framework. The results of this setup, compared to the outcome of setup 1, are illustrated in Figure \ref{fig::noisy_all}. The Root Mean Square Error (RMSE) values for the state components are reported in Table \ref{table:results}. The estimated vehicle trajectory aligns with the ground truth values of the vehicle motion and provides comparable results to the baseline Extended Kalman Filter (EKF) \cite{jai2023_2}. Notably, the EKF provides erroneous yaw angle estimates at the beginning of the experiment, while the FG computes accurate estimates. It is important to consider that the EKF is modeled with knowledge of the vehicle parameters and dynamics, while the FG in this setup only utilizes sensor measurements and knowledge of the static transformations between them. The velocity estimates from the EKF are better due to its utilization of encoder velocity measurements.

\begin{table}[]
\centering
\renewcommand{\arraystretch}{1}
\begin{tabular}{llll}
\hline
\multicolumn{1}{|l}{State}   & \multicolumn{1}{|l|}{Setup 2 FG} & \multicolumn{1}{l|}{Setup 2 EKF}                                        \\ \hline
\multicolumn{1}{|l}{X Position, m}     & \multicolumn{1}{|l|}{0.40} & \multicolumn{1}{l|}{\textbf{0.30}}                       \\ \hline
\multicolumn{1}{|l}{Y Position, m}     & \multicolumn{1}{|l|}{\textbf{0.38}} & \multicolumn{1}{l|}{0.39}                      \\ \hline
\multicolumn{1}{|l}{Yaw Angle, rad}  & \multicolumn{1}{|l|}{\textbf{0.04}} & \multicolumn{1}{l|}{0.06} 
\\ \hline
\multicolumn{1}{|l}{Longitudional Velocity, m/s}   & \multicolumn{1}{|l|}{0.16} &   \multicolumn{1}{l|}{ \textbf{0.147}} 
\\ \hline
&
\end{tabular}
\renewcommand{\arraystretch}{1}
\caption{Root Mean Square Error (RMSE) values for estimated states computed using the FG and EKF algorithm for quadricycle's setup 2}
\label{table:results}
\end{table}

\begin{figure*}
\centering
\includegraphics[width=18cm,height=9cm]{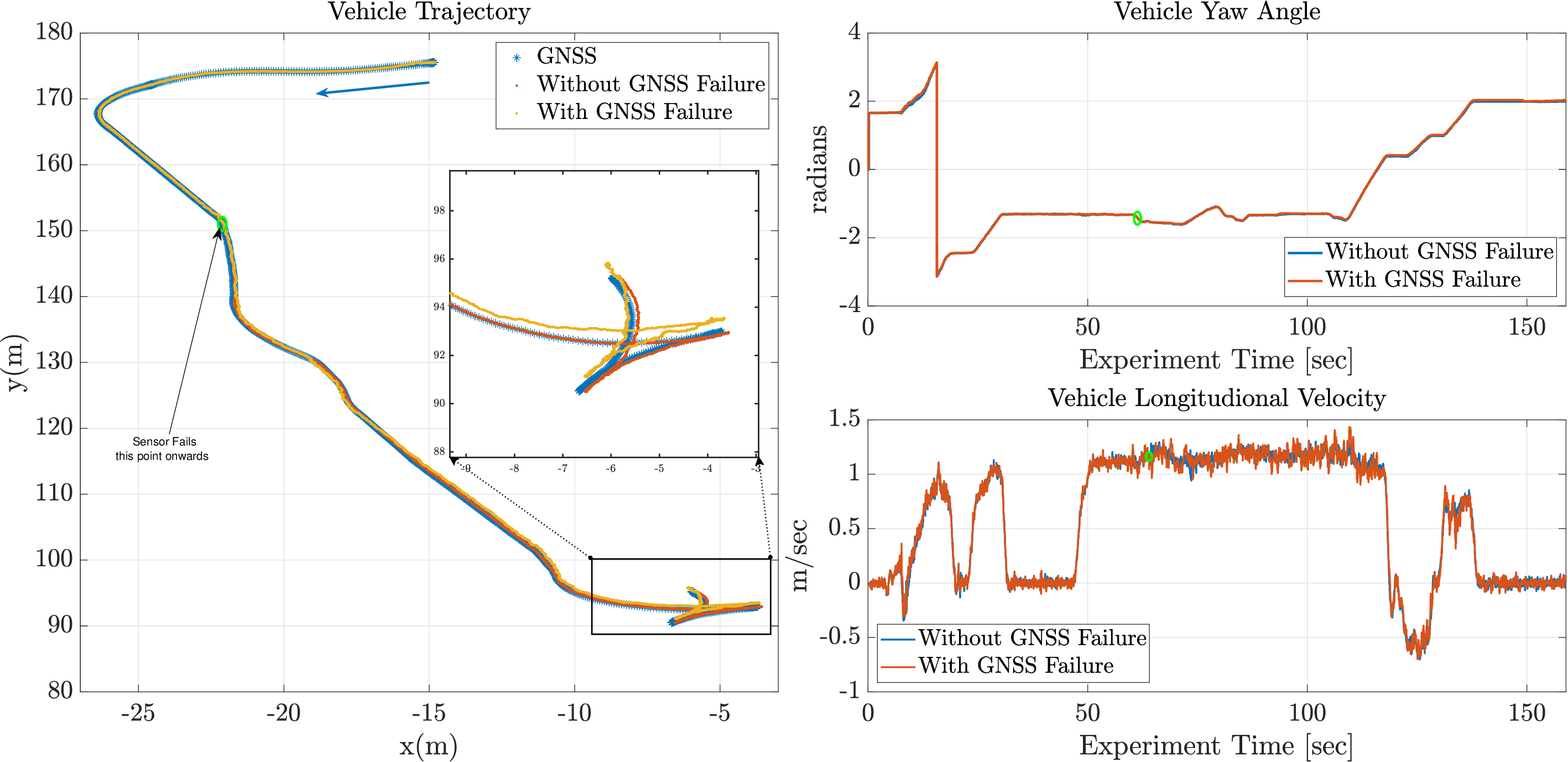}
\caption{State Estimation result for the two studies of the EasyMile Shuttle setup, Setup 3}
\label{fig::easy_all}
\end{figure*}

For EasyMile's setup 3, we conducted two studies. In the first case, all the sensors were in their true state, and no failures were simulated. Since we used an RTK-corrected GNSS sensor, the results of this case study served as the approximate ground truth. To stress-test the algorithm, we simulated a GNSS sensor failure by completely removing its availability to the algorithm starting at $60$ seconds into the experiment until the end. The results of these case studies are illustrated in Figure \ref{fig::easy_all}. Remarkably, the algorithm consistently provided accurate and robust results in both accuracy and robustness. The vehicle trajectory, with and without GNSS failure cases, closely matched the GNSS measurements, and we observed minimal drift in the failure scenario. During the GNSS outage period, we obtained an RMSE of \textbf{0.056 m} and \textbf{0.18 m} in the X and Y coordinates, respectively. The estimation of yaw and velocity remained robust and consistent throughout the experiment, with an RMSE of \textbf{0.020 rad} and \textbf{0.045 m/s}, respectively. After traversing a distance of 77 meters during GNSS failure, a positional divergence of 0.2 meters in the X-axis and 0.6 meters in the Y-axis was observed.\par

\section{Conclusion and Future Work} \label{Conclusion}
In this study, we introduced a state estimator based on Factor Graphs, leveraging the modular nature of this approach and utilizing inputs from the available sensors. Our algorithm was designed to provide reliable and consistent state estimation in different sensor setups, specifically for both quadricycle and shuttle configurations. We also examined the algorithm's performance under simulated scenarios involving GNSS noise and GNSS failure.The proposed algorithm offers several advantages. It adopts a modular architecture that is not dependent on specific vehicle dynamics or sensor setups, making it flexible and adaptable to different scenarios. We utilized the KISS ICP \cite{kissicp} for Lidar Odometry estimation, which is an effective and widely available tool, contributing to the algorithm's accessibility and practical implementation. \par
In our future work, we aim to enhance the algorithm by incorporating additional factors such as Camera Image factors and more complex vehicle dynamics factors. This expansion would allow for further improvements in state estimation accuracy and robustness, especially in cases where detailed vehicle models are available.

\bibliographystyle{IEEEtran}
\bibliography{sample}

\end{document}